\documentclass[journal]{IEEEtran}

\ifCLASSINFOpdf
\else
   \usepackage[dvips]{graphicx}
\fi
\usepackage{url}
\usepackage{graphicx}
\usepackage{tabularx, booktabs, color} 
\usepackage{tabularx} 
\usepackage{booktabs} 
\usepackage{array} 
\usepackage{subcaption} 
\usepackage{multirow}
\usepackage{pifont} 
\newcommand{\cmark}{\ding{51}}
\newcommand{\xmark}{\ding{55}}
\usepackage[table]{xcolor} 
\usepackage{colortbl} 
\usepackage{booktabs} 
\usepackage{amsmath}
\definecolor{myred}{rgb}{0.98,0.81,0.81} 
\definecolor{myblue}{rgb}{0.67,0.81,0.91} 
\usepackage{amsfonts} 

\begin{document}

\title{CMT: Cross Modulation Transformer with Hybrid Loss for  Pansharpening }

\author{Wen-Jie Shu,  Hong-Xia Dou, Rui Wen, Xiao Wu and Liang-Jian Deng, \IEEEmembership{Senior Member, IEEE}
\thanks{The research is supported by NSFC (No. 12271083), and National Key
Research and Development Program of China (No. 2020YFA0714001).}
\thanks{Wen-Jie Shu is with the School of Optoelectronic Science and Engineering, University of Electronic Science and Technology of China, Chengdu
611731, China (e-mail: wenjieshu2003@gmail.com). Hong-Xia Dou is with the School of Science, Xihua University, Chengdu 610039, China (e-mail: hongxia.dou@mail.xhu.edu.cn). Rui Wen, Xiao Wu and Liang-Jian Deng are with the School of
Mathematical Sciences, University of Electronic Science and Technology of China, Chengdu 611731, China (e-mail: wenrui202102@163.com; wxwsx1997@gmail.com; liangjian.deng@uestc.edu.cn). }}

\markboth{Journal of \LaTeX\ Class Files, Vol. 14, No. 8, Aprial 2024}
{Shell \MakeLowercase{\textit{et al.}}: Bare Demo of IEEEtran.cls for IEEE Journals}
\maketitle

\begin{abstract}
Pansharpening aims to enhance remote sensing image (RSI) quality by merging high-resolution panchromatic (PAN) with  multispectral (MS) images.  
However, prior techniques struggled to optimally fuse PAN and MS images for enhanced spatial and spectral information, due to a lack of a  systematic framework capable of effectively coordinating their individual strengths.
In response, we present the Cross Modulation Transformer (CMT), a pioneering method that modifies the attention mechanism. This approach utilizes a robust modulation technique from signal processing, integrating it into the attention mechanism's calculations. It dynamically tunes the weights of the carrier's value (V) matrix according to the modulator's features, thus resolving historical challenges and achieving a seamless integration of spatial and spectral attributes.
Furthermore, considering that RSI exhibit large-scale features and edge details along with local textures, we crafted a hybrid loss function that combines Fourier and wavelet transforms to effectively capture these characteristics, thereby enhancing both spatial and spectral accuracy in pansharpening.
Extensive experiments demonstrate our framework's superior performance over existing state-of-the-art methods. The code will be publicly available to encourage further research.
\end{abstract}


\IEEEpeerreviewmaketitle

\section{Introduction}
\IEEEPARstart{G}{iven} the inherent constraints of remote sensing technology, obtaining MS images with high spatial resolution directly from satellites is a significant challenge. As a solution, pansharpening has become a crucial technique. It merges low-resolution MS (LRMS) images with high-resolution PAN images to produce high-spatial-resolution MS (HRMS) images with superior spatial detail. This fusion technique effectively navigates around the limitations of sensor technology, offering invaluable data for RSI analysis.
\begin{figure}[tb]
  \centering
  \includegraphics[height=5cm]{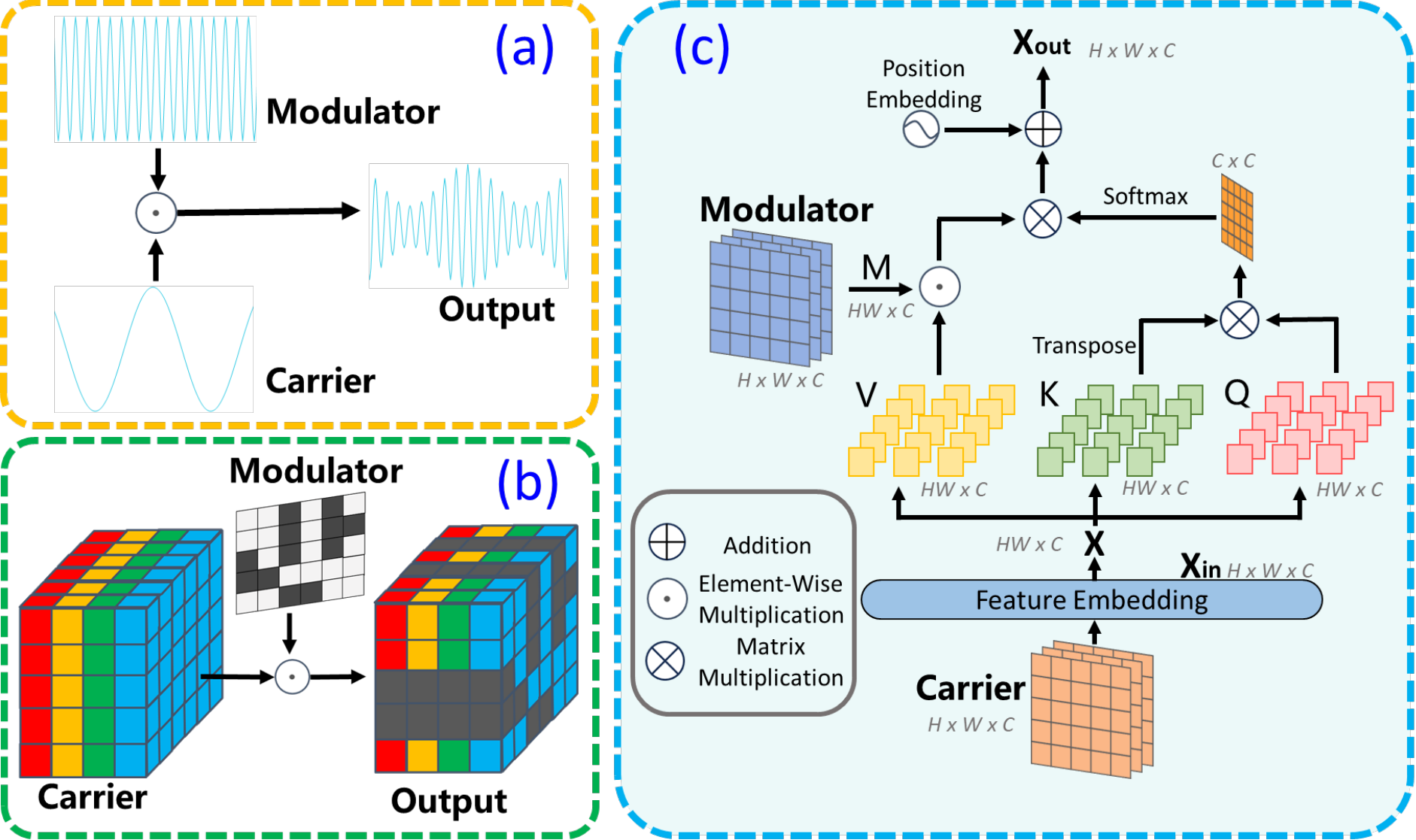}
  \caption{ 
   (a) The modulation process in the field of communication. (b) The modulation process in coded aperture snapshot spectral imaging (CASSI).  (c) Our Cross Modulation Multi-head Self-Attention (CM-MSA) modulation process.  
  }
  \label{fig:example}
\end{figure}

Deep learning breakthroughs, led by Convolutional Neural Networks (CNNs), have significantly advanced the field of pansharpening \cite{chen2022spanconv}, \cite{jin2021bam}, \cite{zhang2022triple}. CNNs have shown exceptional prowess in pansharpening, skillfully extracting and combining intricate features from various images to enhance both spatial and spectral quality. Furthermore, since their introduction, Transformers \cite{vaswani2017attention} have revolutionized numerous fields, including pansharpening \cite{hu2022fusformer}, \cite{zhang2023p2sharpen}, \cite{visiontransformerpan} by their unparalleled ability to model long-range dependencies using self-attention mechanisms. This capability gives them a significant edge in effectively blending spatial and spectral features, outperforming traditional CNN-based approaches by capturing the complex dynamics between different types of images.

However, the field of pansharpening still faces several challenges. 
\textbf{Firstly}, Hyperspectral Images (HSI) show spatial sparsity and spectral self-similarity, complicating spatial dependency modeling and highlighting the importance of prioritizing inter-spectral over spatial correlations in Transformers. \textbf{Secondly}, RSI are marked by their rich spectral content, complex surface textures and edge details, posing challenges for fusion and necessitating tailored approaches for effective integration. \textbf{Lastly}, the potential of attention structures, specifically tailored for pansharpening within the Transformer framework, warrants further exploration.
Current integration methods of PAN and LRMS images within frameworks like PanFormer \cite{zhou2022panformer} and Hyperformer \cite{bandara2022hypertransformer} mainly involve linear projection of images into tokens for merging via attention computations. While these methods are efficient, they don't fully exploit the intricate relationship between the spatial resolution of PAN and the spectral diversity of LRMS images.

\begin{figure}[tb]
  \centering
  \includegraphics[height=4.5cm]{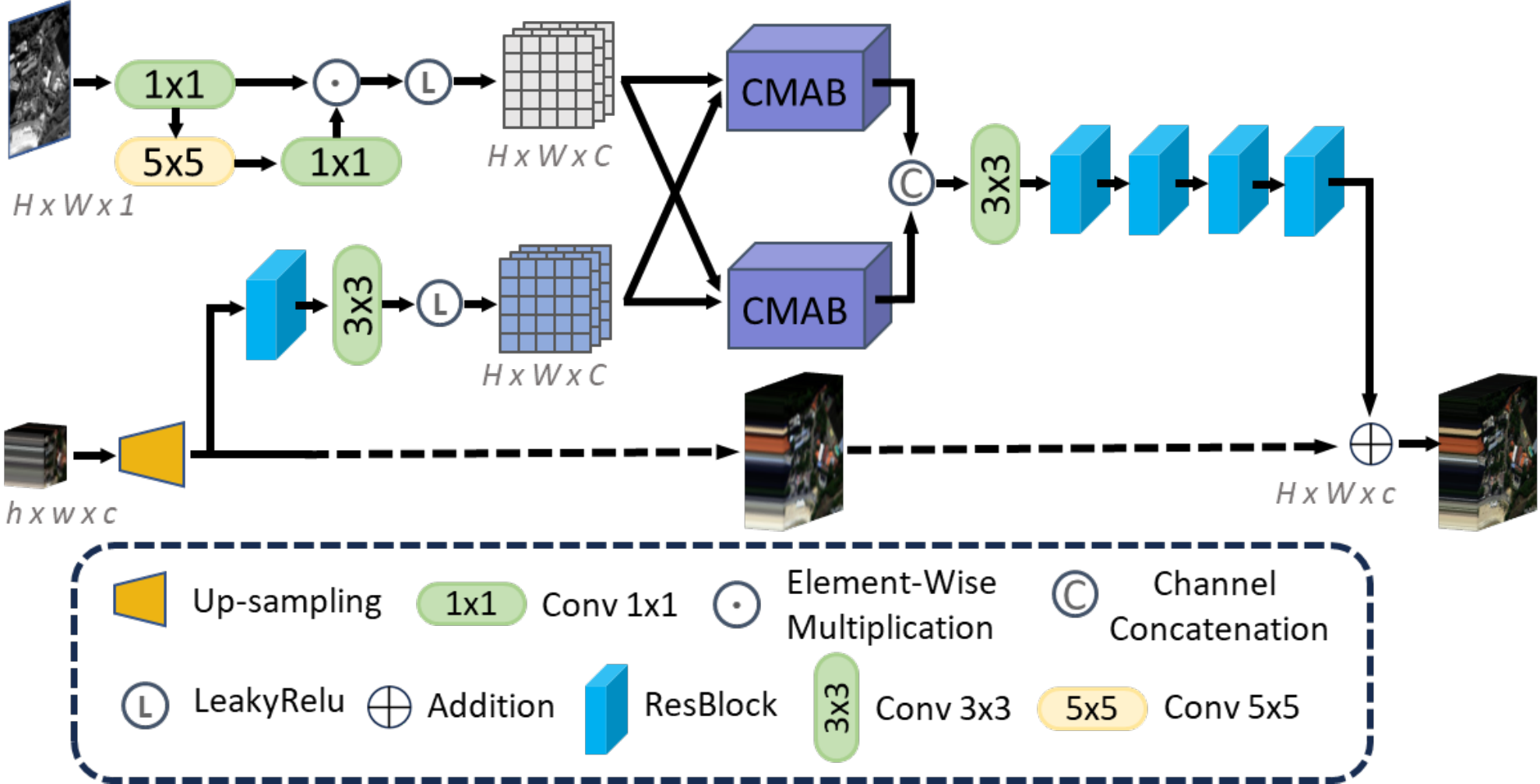}
  \caption{ 
   Overall structure of the proposed method. 
  }
  \label{fig:example}
\end{figure}

In response to these challenges, we have developed innovative solutions, notably the CMT and a hybrid loss function, to elevate the pansharpening workflow.
\textbf{Firstly},  as illustrated in Fig. 1 (c), to harness spectral features for LRMS images and spatial correlation for PAN images, we independently compute the attention block for each spectral and spatial channel ensuring a targeted processing of the distinct characteristics inherent. \textbf{Secondly}, to address the nuanced complexities of RSI, characterized by their rich spectral content and complex surface textures, we've implemented a hybrid loss function that combines Fourier and wavelet transforms. This approach utilizes Fourier transforms for identifying widespread features and wavelet transforms for enhancing local texture details. Together, they effectively improve spatial detail and maintain spectral fidelity in the pansharpening process. 
\textbf{Lastly}, our approach innovatively blends PAN's spatial details with LRMS's spectral data by applying advanced modulation techniques to the pansharpening process. 
Specifically, rather than merely concatenating features, our approach leverages the features of the modulator to dynamically modulate the carrier, altering the weights of the Transformer's value (V) matrix, which achieves a more sophisticated fusion of features.
Our approach, inspired by signal processing techniques depicted in Fig. 1 (a), uses modulation to seamlessly integrate high and low-frequency signals, significantly improving signal fidelity and richness. In parallel, CASSI systems, illustrated in Fig. 1 (b), compress images by modulating high frequencies with masks, akin to our deep learning methods for image enhancement like MST by Cai et al \cite{cai2022mask}, which also employ masks for recovery.
 In conclusion, the CMT framework significantly advances pansharpening through the following contributions:
\begin{enumerate}
    \item Our Cross Modulation module within the CMT framework significantly enhances the fusion of PAN and LRMS images through a novel modulation technique.
    \item We introduce a pioneering hybrid loss function that combines Fourier and wavelet transforms, which is the first attempt in the field of pansharpening to the best of our knowledge.
    \item The CMT framework delivers outstanding results on benchmark datasets, establishing a new benchmark for pansharpening performance.
\end{enumerate}

\begin{figure}[tb]
  \centering
  \includegraphics[height=6cm]{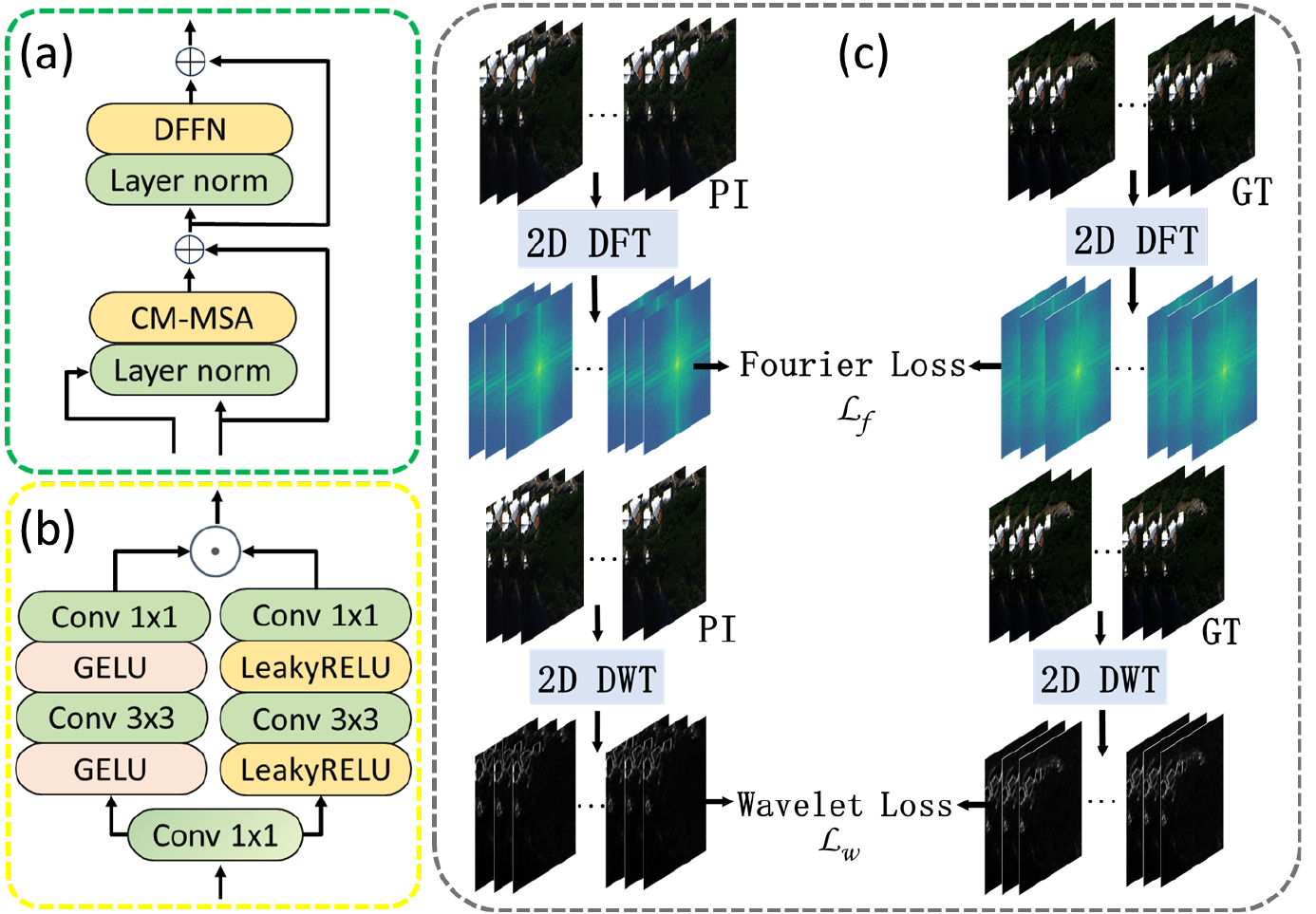}
  \caption{ 
   (a) The CMAB module consists of a Double Feed-Forward Network (DFFN), a CM-MSA module, and two layers of normalization. (b) Components of the DFFN. (c) The hybrid loss between Predicted Images ($PI$) and ground truth ($GT$), which employs both the 2D Discrete Fourier Transform (DFT) and the 2D Discrete Wavelet Transform (DWT).}
  
  \label{fig:example}
\end{figure}

\section{Method}

\subsection{Overall Architecture}
The CMT architecture, depicted in Fig. 2, is structured into three primary phases: feature extraction, modulation, and feature aggregation. In the feature extraction phase, distinct extractors are designed to capture the unique characteristics of PAN and LRMS images. During the modulation phase, the Cross Modulation Attention Block (CMAB) modulator separately modulates PAN images to enrich LRMS images and vice versa, allowing for an effective blend of spatial and spectral information. Finally, to aggregate feature, we use a \(3 \times 3\) convolutional kernel and four ResNet blocks simply.

Initially, LRMS images are upscaled to dimensions $H \times W \times c$. Then,  PAN and LRMS images are processed through convolutional and ResNet blocks to extract local and global spatial details, resulting in feature sets \(F_{\text{pan}}\) and \(F_{\text{ms}}\), respectively, with channel expansion to \(C=32\) in implementation. 


The extracted features \(F_{\text{pan}}\) and \(F_{\text{ms}}\) are then modulated within the CMAB module, as shown in Fig. 3 (a). 
This module comprises a CM-MSA, two layers of normalization and a DFFN, employing varied activation functions for enhanced modulation. Fig. 3 (b) illustrates the specifics of the DFFN.
Post-modulation, the features are concatenated and merged through a \(3 \times 3\) convolution and four ResNet blocks, allowing for further integration of spatial and spectral information. 

Ultimately, the aggregated features are combined with the upsampled LRMS images to produce the HRMS.


\subsection{Modulation Approach}
In signal processing, Double Side-Band modulation (DSB) \cite{hartley1928transmission}, \cite{modulation} stands as a classic and effective technique wherein the amplitude of a carrier wave is varied in accordance with the instantaneous value of the message signal, thus encoding information within the carrier. The mathematical expression for an DSB signal can be accurately given by:
\begin{equation}
s(t) = m(t) \cdot A_c \cdot \cos(2\pi f_c t),
\end{equation}
where $s(t)$ denotes the modulated signal, $A_c$ represents the constant amplitude of the carrier wave, $f_c$ is the carrier frequency, and $m(t)$ embodies the message signal intended for transmission. 
This foundational concept of modulation underscores its robust capability to encode and transmit complex information efficiently and accurately across different mediums.
In a similar vein, depicted in Fig. 1 (c), our pansharpening framework leverages a cross modulation paradigm where high-resolution spatial features and spectral details are mutually modulated. This bilateral modulation mechanism considerably amplifies the model’s proficiency in capturing and amalgamating multi-dimensional information, marking a significant leap from traditional methods.

Firstly, for the carrier, the input feature \(X_{\text{in}}\) $\in \mathbb{R}^{H \times W \times C}$ is reshaped into tokens $X \in \mathbb{R}^{HW \times C}$. A multi-head attention mechanism is employed to improve generalization and capture multi-dimensional information, splitting $X$ into $k$ heads:
\begin{equation}
    X = [X_1, X_2, \ldots, X_k],
\end{equation}
where $X_i \in \mathbb{R}^{HW \times d_k}$, $d_k = \frac{C}{k}$, and $i = 1, 2, \ldots, k$.

Each $X_i$ is then linearly projected into queries $Q_i$, keys $K_i$, and values $V_i$ using the following equations:
\begin{equation}
    Q_i = X_iW^Q_i, \quad K_i = X_iW^K_i, \quad V_i = X_iW^V_i,
\end{equation}
where $W^Q_i$, $W^K_i$, and $W^V_i \in \mathbb{R}^{d_k \times C}$ are learnable parameters, and $i = 1, 2, \ldots, k$.

For the modulator, aligned with the carrier, the input feature is reshaped into tokens $M \in \mathbb{R}^{HW \times C}$ and split into $k$ heads. Then, the modulation is integrated into the self-attention calculation by element-wise multiplying $M_i$ with $V_i$:
\begin{equation}
    M = [M_1, M_2, \ldots, M_k],  \quad  V'_i = M_i \odot V_i 
\end{equation}
where $M_i \in \mathbb{R}^{HW \times d_k}$, $i = 1, 2, \ldots, k$.



For a single head, the modulation-attention computation is:
\begin{equation}
\begin{split}
    \text{MA}(Q_i, K_i, V'_i) 
    &= V'_i \cdot \text{softmax}\left(\frac{K_i^T Q_i}{\alpha_i}\right),
\end{split}
\end{equation}
where MA is modulation attention and $\alpha_i \in \mathbb{R}$ is a learnable parameter that adaptively scales the matrix multiplication, enhancing the model's ability to adjust attention weights dynamically.

The results from multi-heads are concatenated together, and with the addition of position encoding, the final output $X_{\text{out}}$ is derived as follows:
\begin{equation}
    X_{\text{out}} = \text{Fc}(\text{Concatenate}(\text{MA}_1, \text{MA}_2, \ldots, \text{MA}_k)) + F_p,
\end{equation}
where $\text{Fc}$ denotes a fully connected layer, and $F_p$ represents the position encoding.

\begin{table*}[ht]
\centering
\caption{Quantitative results on 20 reduced-resolution and 20 full-resolution samples of GF2. (\textcolor{red}{Red}: best; \textcolor{blue}{Blue}: second best).}
\label{tab:metrics_comparison}
\renewcommand{\arraystretch}{1.0} 
\begin{tabular}{m{2cm}cccccc}
\toprule
\multirow{2}{*}{\centering \textbf{Method}} & \multicolumn{3}{c}{\textbf{Reduced-Resolution Metrics}} & \multicolumn{3}{c}{\textbf{Full-Resolution Metrics}} \\
\cmidrule(lr){2-4} \cmidrule(lr){5-7}
& \textbf{SAM↓} & \textbf{ERGAS↓} & \textbf{Q4↑} & \textbf{\(D_{\lambda}\)↓} & \textbf{\(D_{s}\)↓} & \textbf{HQNR↑} \\ 
\midrule
\centering \textbf{PNN}\cite{PNN} & 1.048 $\pm$ 0.226 & 1.057 $\pm$ 0.235 & 0.960 $\pm$ 0.010 & 0.0317 $\pm$ 0.0286 & 0.0943 $\pm$ 0.0224 & 0.877 $\pm$ 0.036 \\ 
\centering \textbf{PanNet}\cite{PanNet}  & 0.997 $\pm$ 0.212 & 0.919 $\pm$ 0.191 & 0.967 $\pm$ 0.010 & \textcolor{red}{0.0179} $\pm$ 0.0110 & 0.0799 $\pm$ 0.0178 & 0.904 $\pm$ 0.020 \\
\centering \textbf{DiCNN}\cite{DiCNN}  & 1.052 $\pm$ 0.231 & 1.081 $\pm$ 0.254 & 0.959 $\pm$ 0.010 & 0.0369 $\pm$ 0.0132 & 0.0992 $\pm$ 0.0131 & 0.868 $\pm$ 0.016 \\ 
\centering \textbf{FusionNet}\cite{FusionNet}  & 0.973 $\pm$ 0.212 & 0.988 $\pm$ 0.222 & 0.964 $\pm$ 0.009 & 0.0350 $\pm$ 0.0124 & 0.1013 $\pm$ 0.0134 & 0.867 $\pm$ 0.018  \\ 
\centering \textbf{DCFNet}\cite{DCFNet} & 0.872 $\pm$ 0.169 & 0.784 $\pm$ 0.146 & 0.974 $\pm$ 0.009 & 0.0240 $\pm$ 0.0115 & \textcolor{blue}{0.0659} $\pm$ 0.0096 & \textcolor{blue}{0.912} $\pm$ 0.012  \\ 
\centering \textbf{MMNet}\cite{MMNet}  & 0.993 $\pm$ 0.141 & 0.777 $\pm$ 0.134 & 0.969 $\pm$ 0.020 & 0.0443 $\pm$ 0.0298 & 0.1033 $\pm$ 0.0129 & 0.857 $\pm$ 0.027  \\ 
\centering \textbf{LAGConv}\cite{LAGConv}  & \textcolor{blue}{0.786} $\pm$ 0.148 & 0.687 $\pm$ 0.113 & 0.980 $\pm$ 0.009 & 0.0284 $\pm$ 0.0130 & 0.0792 $\pm$ 0.0136 & 0.895 $\pm$ 0.020  \\ 
\centering \textbf{HMPNet}\cite{HMPNet}  & 0.803 $\pm$ 0.156 & \textcolor{red}{0.564} $\pm$ 0.099 & \textcolor{blue}{0.981} $\pm$ 0.030 & 0.0819 $\pm$ 0.0499 & 0.1146 $\pm$ 0.0126 & 0.813 $\pm$ 0.049  \\ 
\centering \textbf{Proposed}  & \textcolor{red}{0.722} $\pm$ 0.136 & \textcolor{blue}{0.624} $\pm$ 0.107 & \textcolor{red}{0.992} $\pm$ 0.001 & \textcolor{blue}{0.0202} $\pm$ 0.0103 & \textcolor{red}{0.0338} $\pm$ 0.0086 & \textcolor{red}{0.947} $\pm$ 0.012  \\ \hline

\end{tabular}
\end{table*}

\begin{figure*}[tb]
  \centering
  \includegraphics[height=4.5cm]{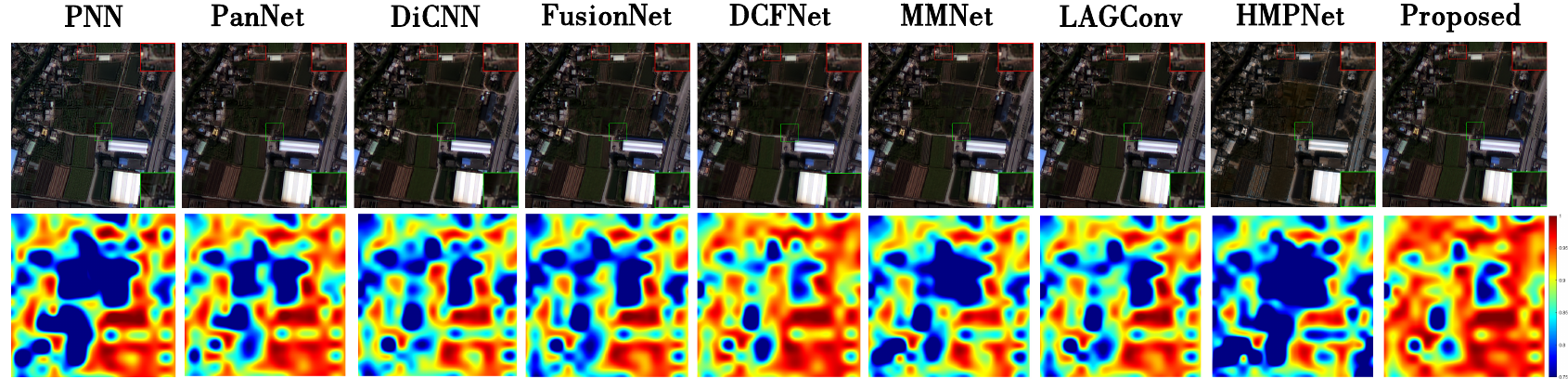}
  \caption{ 
   Qualitative result comparison between representative methods on the GF2 reduced-resolution dataset. The first row presents RGB outputs, while the second row gives the corresponding QNR maps.
  }
  \label{fig:example}
\end{figure*}

\subsection{Loss Function}

To enhance the resolution and quality of RSI, as shown in Fig. 3 (c), our method improves upon traditional fusion techniques by leveraging both Fourier and wavelet transforms. Fourier transforms \cite{paley1934fourier} are essential for mapping images into the frequency domain, capturing widespread environmental features. Besides, wavelet transforms \cite{wavelet} excel in delineating images across multiple scales, adeptly enhancing local textures and detailing.

The Fourier transform loss function is defined as:
\begin{equation}
L_{\text{Fourier}} = \frac{1}{M} \sum_{i=1}^{M} \| \mathcal{F}(PI_i) - \mathcal{F}(GT_i) \|_1,
\end{equation}
which applies the L1 loss to the difference in Fourier-transformed Predicted Images ($PI_i$) and ground truth images ($GT_i$), averaged over all $M$ training samples.

The loss function for the wavelet transform is defined as:
\begin{equation}
L_{\text{Wavelet}} = \frac{1}{M} \sum_{i=1}^{M}  D_{j,c}(i),
\end{equation}
where $D_{j,c}(i) = \| \mathcal{W}{j,c}(PI_i) - \mathcal{W}{j,c}(GT_i) \|_1$ and $\mathcal{W}{j,c}$ captures wavelet coefficients at scale $j$ and orientation $c$, reflecting local variations and textures, averaged across all training instances.


Combining these components, our comprehensive loss function integrates spatial, frequency, and wavelet domain losses:
\begin{equation}
L_{\text{total}} = L_{\text{spa}} + \lambda_{1} L_{\text{Fourier}} + \lambda_{2} L_{\text{Wavelet}}
\end{equation}
where $L_{\text{spa}}$ is the L1 loss in the spatial domain, $\lambda_{1}$ and $\lambda_{2}$ are set to 0.7 and 0.2 to balance the contributions of the different loss components as weighting coefficients in implementation.

\section{Experiment}
\subsection{Datasets and Implementation Details}

To validate our approach, we construct datasets following Wald’s protocol \cite{deng2020detail}, \cite{wald1997fusion} on data collected from the WorldView-3 (WV3) and GaoFen-2 (GF2) satellites. Our datasets and data processing methods are downloaded from the PanCollection repository \cite{deng2022machine}. The datasets consist of images cropped from entire remote sensing images, divided into training and testing sets. The training set comprises PAN/LRMS/GT image pairs obtained by downsampling simulation, with dimensions of 64×64, 16×16×C and 64×64×C. Besides, we evaluated our method on the commonly used metrics in the field of pansharpening, including  SAM \cite{SAM},  ERGAS \cite{ERGAS}  and Q8 \cite{Q8} for reduced-resolution dataset, \(Ds\), \(D_{\lambda}\) and HQNR \cite{QNR} for full-resolution dataset. The CMT was trained with an initial learning rate of 0.001, for 400 epochs, and a batch size of 32, using Adam optimizer \cite{adam} with learning rate halved every 100 epochs. As for other DL-based methods, we utilize the default settings in related papers or codes to train the networks.

\begin{table}[ht]
\centering
\caption{Quantitative results of DL-based methods on 20 reduced-resolution samples acquired by WV3. (\textcolor{red}{Red}: best; \textcolor{blue}{Blue}: second best).}
\label{tab:WV3_Reduced}
\begin{tabular}{lccc}
\toprule
\textbf{Method} & \textbf{Q8↑} & \textbf{SAM↓} & \textbf{ERGAS↓} \\
\midrule
PNN & 0.893 $\pm$ 0.092 & 3.677 $\pm$ 0.762 & 2.680 $\pm$ 0.647 \\
PanNet & 0.891 $\pm$ 0.093 & 3.613 $\pm$ 0.766 & 2.664 $\pm$ 0.688 \\
DiCNN & 0.900 $\pm$ 0.087 & 3.592 $\pm$ 0.762 & 2.672 $\pm$ 0.662 \\
FusionNet & 0.904 $\pm$ 0.090 & 3.324 $\pm$ 0.698 & 2.465 $\pm$ 0.644 \\
MMNet & 0.915 $\pm$ 0.086 & 3.084 $\pm$ 0.640 & 2.343 $\pm$ 0.626 \\
LAGConv & 0.910 $\pm$ 0.091 & 3.103 $\pm$ 0.558 & 2.292 $\pm$ 0.607 \\
HMPNet & \textcolor{blue}{0.916} $\pm$ 0.087 & \textcolor{blue}{3.063} $\pm$ 0.577 & \textcolor{blue}{2.229} $\pm$ 0.545 \\
Proposed & \textcolor{red}{0.917} $\pm$ 0.086 & \textcolor{red}{3.001} $\pm$ 0.610 & \textcolor{red}{2.201} $\pm$ 0.522 \\
\bottomrule
\end{tabular}
\end{table}

\begin{table}[ht]
\centering
\caption{Ablation study on loss components. (\textcolor{red}{Red}: best; \textcolor{blue}{Blue}: second best).}
\label{tab:ablation_study}
\begin{tabular}{ccc cccccc}
\toprule
\multicolumn{3}{c}{Loss Components} & \multicolumn{3}{c}{Metrics} \\
\cmidrule(lr){1-3} \cmidrule(lr){4-6}
L$_{\text{spa}}$ & L$_{\text{wave}}$ & L$_{\text{fft}}$  & \textbf{Q8↑} & \textbf{SAM↓} & \textbf{ERGAS↓} \\
\midrule
\cmark & \xmark & \xmark  & 0.912±0.086 & 3.043±0.618 & 2.238±0.530 \\
\cmark & \cmark & \xmark  & 0.913±0.087 & 3.033±0.612 & 2.223±0.523 \\
\cmark & \xmark & \cmark  & \textcolor{blue}{0.916}±0.086 & \textcolor{blue}{3.006}±0.609 & \textcolor{blue}{2.208}±0.521 \\
\cmark & \cmark & \cmark  & \textcolor{red}{0.917}±0.086 & \textcolor{red}{3.001}±0.610 & \textcolor{red}{2.201}±0.522 \\
\bottomrule
\end{tabular}
\end{table}

\begin{table}[ht]
\centering
\caption{Ablation study on Modulation Approach. (\textcolor{red}{Red}: best; \textcolor{blue}{Blue}: second best).}
\label{tab:performance_comparison}
\begin{tabular}{lccc}
\toprule
\textbf{Method} & \textbf{\(D_{\lambda}\)↓} & \textbf{\(Ds\)↓} & \textbf{HQNR↑} \\
\midrule
V1 & \textcolor{blue}{0.0210}±0.0074 & 0.0364±0.0125 & \textcolor{blue}{0.9435}±0.0180  \\
V2 & 0.0249±0.0123 & \textcolor{blue}{0.0355}±0.0132 & 0.9406±0.0187  \\
V3 & 0.0234±0.0079 & 0.0388±0.0155 & 0.9389±0.0210 \\
CMT& \textcolor{red}{0.0201}±0.0074 & \textcolor{red}{0.0344}±0.0135 & \textcolor{red}{0.9463}±0.0188  \\
\bottomrule
\end{tabular}
\end{table}

\subsection{Results}
The performance of the proposed CMT method is show cased through extensive evaluations on GF2 datasets. TABLE I present a comprehensive comparison of CMT with various state-of-the-art methods on the GF2 dataset. The quantitative results show our method consistently surpasses existing approaches in all metrics. The visual comparison results are provided in Fig. 3. TABLE II present the results on the WV3 datasets, and the proposed method obtains the best average results on all quality indexes. 
\subsection{Ablation Experiment}
\textbf{Ablation on Hybrid Loss.} We compare the outcomes after training with different loss functions. The results are displayed in TABLE III. This comparison aims to ascertain the impact of each loss component on the overall performance of the model. We perform experiments on 20 reduced-resolution samples acquired by WV3 satellite.

\textbf{Ablation on Modulation Approach.} To validate the effectiveness of our method, we create three variants of the CMT. In the first variant (V1), modulation is omitted, taining only the transformer to evaluate its baseline feature integration performance. The second variant (V2) involves exclusively using the PAN image to modulate the MS features.
In the third variant (V3), we solely use LRMS modulation of the PAN image.
We perform experiments on 20 full-resolution samples acquired by the WV3 satellite. 
The results in TABLE IV show CMT has the best overall performance, proving our method's efficacy.


\section{Conclusion}

In this study, we introduce the CMT, a novel pansharpening method that synergistically merges PAN and LRMS images. Central to CMT is the application of signal modulation, innovatively incorporated into a Transformer-based architecture. This allows for precise modulation of the Transformer's value (V) matrix, facilitating a superior integration of spatial detail and spectral depth. Our method is characterized by its unique CM-SMA modulation technique and a bespoke hybrid loss function that blends Fourier and wavelet transforms. This loss function adeptly captures both global patterns and local textures, thereby enhancing spatial resolution while maintaining spectral fidelity. The versatility of CMT suggests its applicability beyond pansharpening, offering promising enhancements in various fields that require intricate and spectrally accurate image fusion.

\bibliographystyle{plain}
\bibliography{bare_jrnl-docx}

\end{document}